\documentclass[10pt,twocolumn,letterpaper]{article}

\usepackage{cvpr}
\usepackage{times}
\usepackage{epsfig}
\usepackage{graphicx}
\usepackage{amsmath}
\usepackage{amssymb}

\usepackage{booktabs} 
\usepackage{bm}
\usepackage{bbding}
\usepackage{array}
\usepackage{caption}

\cvprfinalcopy 

\def\cvprPaperID{****} 

\begin{document}

\title{Temporal Convolution Based Action Proposal: Submission to ActivityNet 2017}

\author{Tianwei Lin$^1$, Xu Zhao$^1$\thanks{Corresponding author.} , Zheng Shou$^2$\\
$^1$Computer Vision Laboratory, Shanghai Jiao Tong University, China. $^2$Columbia University, USA\\
{\tt\small \{wzmsltw,~zhaoxu\}@sjtu.edu.cn, zs2262@columbia.edu}
}

\maketitle

\begin{abstract}

In this notebook paper, we describe our approach in the submission to  the {\bf temporal action proposal (task 3)} and {\bf temporal action localization  (task 4)} of ActivityNet Challenge hosted at CVPR 2017. 
Since the accuracy in action classification task is already very high (nearly $90\%$ in ActivityNet dataset), we believe that the main bottleneck for temporal action localization is the quality of action proposals. Therefore, we mainly focus on the temporal action proposal task and propose a new proposal model based on temporal convolutional network. 
Our approach achieves the state-of-the-art performances on both temporal action proposal task and temporal action localization task.

\end{abstract}

\section{Introduction}

Action recognition and temporal action localization are both important branches of video content analysis. The temporal action localization or detection task aims to detect action instances in untrimmed video, containing categories and temporal boundaries of action instances. 

Temporal action localization task can be divided into two main parts. (1) Temporal action proposal, which means we need generate some temporal boundaries of action instances without classifying their categories. (2) Action recognition (or we can say action classification), in this part we need to decide the categories of temporal action proposals. Most of previous works \cite{shou2016action,xiong2017pursuit} address these two parts separately.  There are also works \cite{escorcia2016daps,fast_temporal_activity_cvpr16} focusing on temporal action proposal.   For action recognition, there are already many algorithms \cite{wang2016temporal,feichtenhofer2016convolutional} with great performance. However, the localization accuracy (mean average precision) is still very low in multiple benchmarks such as THUMOS'14 \cite{jiang2014thumos} and ActivityNet \cite{caba2015activitynet}, comparing with the situation in object localization. We think the main constraint on accuracy of temporal action localization  is the quality of action proposals. Therefore, we mainly focus on the temporal action proposal task in this challenge and our high quality proposals also lead to state-of-the-art performance in temporal action localization task.


\section{Our Approach}

\begin{figure*}
\centering

\begin{minipage}[b]{1.0\linewidth}
  \centering
  \centerline{\includegraphics[width=17cm]{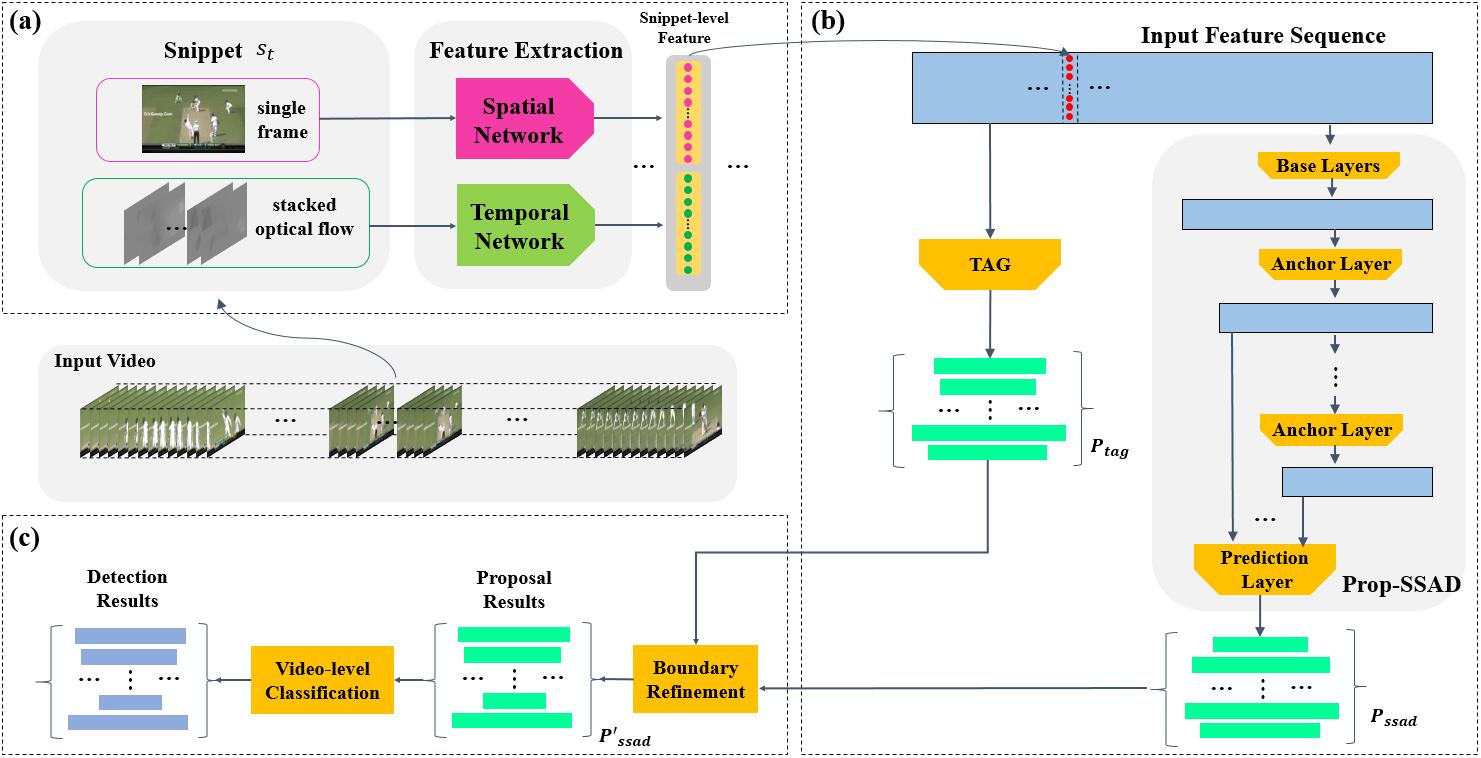}}
  \medskip
\end{minipage}

\caption{The framework of our approach. (a) Two-stream networks are used to extract snippet-level features. (b) Prop-SSAD model and TAG method are used for proposal generation separately. (c) Proposals generated by TAG are used for refining the boundaries of proposals generated by Prop-SSAD model. We use video-level action classification result as the category of temporal action proposals to get temporal action localization result.
}
\label{approach}

\end{figure*}

The framework of our approach is shown in Fig \ref{approach}. In this section, we  introduce each part of the framework, which consists of feature extraction, temporal action proposal and temporal action localization.

\subsection{Feature Extraction}

The first step of our framework is feature extraction. We extract two-stream features in a similar way described in \cite{gao2017cascaded}. We adopt two-stream network \cite{xiong2016cuhk} which is pre-trained on ActivityNet v1.3 training set. First we segment video  into 16-frames snippets without overlap. In each snippet, we use spatial network to extract appearance feature with central frame, and we use the output of ``Flatten-673" layer in ResNet network as feature. For motion feature, we compute optical flows using 6 consecutive frames around the center frame of a snippet, then these optical flows are used for extracting motion feature with temporal network, where the output of ``global-pool" layer in BN-Inception network is used as feature. Then, we concatenate appearance and motion feature to form the snippet-level features, which are 3072-dimensional vectors. So after feature extraction, we can transfer a video into a sequence of snippet-level feature vectors. Finally, we resize the feature sequence to new length 256 by linear interpolation.

Since we only use two-stream network  trained on ActivityNet v1.3 training set to extract features, there is no external data used in our approach.

\subsection{Temporal Action Proposal}

{\bf Prop-SSAD.} In our previous work \cite{ssad} \footnote{This paper can be found at: https://wzmsltw.github.io/}, we design a model called {\bf Single Shot Action Detector  (SSAD)} network which simultaneously conducts temporal action proposal and recognition. A core idea of SSAD is applying anchor mechanism to temporal action localization task based on temporal convolutional layers, which is similar with YOLO \cite{redmon2016you} and SSD \cite{liu2016ssd} network for object localization task. In detail, we associate multiple temporal anchor instances with multi-scale temporal feature maps, then use temporal convolutional layers to predict information of anchor instances, including action categories, overlap score and location offsets. So SSAD can directly detect temporal action instances using feature sequence of untrimmed video.

In this challenge, we use SSAD network to make temporal action proposal without action recognition and we call it {\bf Prop-SSAD}. The main differences of network configuration between Prop-SSAD and SSAD are listed below.

\begin{itemize}  
\item Type of input features. In SSAD, we use two-stream network and C3D network to extract feature of video; in Prop-SSAD, only two-stream networks are used.
\item Number of anchor layers. In SSAD, we only associate temporal anchors with 3 temporal feature maps using anchor layers with length 4, 8 and 16; in  Prop-SSAD, we associate temporal anchors with 7 temporal feature maps with length 1, 2, 4, 8, 16, 32, 64.
\item Loss function. In SSAD, we use classification, overlap and location loss jointly to train network; in Prop-SSAD, only overlap loss is used.
\end{itemize}  

{\bf TAG \cite{xiong2017pursuit}.} We also implement {\bf Temporal Actionness Grouping (TAG)} method to generate temporal action proposals, which is proposed in \cite{xiong2017pursuit}. Since the code of TAG is not released yet, we implement TAG by ourselves. First we train a multi-layer perceptron (MLP) model with one hidden layer to predict the actionness score for each snippet, then we use grouping method described in \cite{xiong2017pursuit} with multiple threshold to generate temporal action proposals. Proposals generated by TAG are used for refining the proposals' boundaries generated by Prop-SSAD.

{\bf Boundaries Refinement.} Given feature vector sequence of a video, we can get  temporal action proposals set $P_{ssad}$ using Prop-SSAD and temporal action proposals set $P_{tag}$ using TAG. For each proposal $p_t$ in $P_{tag}$, we calculate its IoU with all proposals in $P_{ssad}$. If the maximum IoU is higher than threshold 0.75, we replace the boundaries of corresponding proposal $p_s$ in $P_{ssad}$ with boundaries of $p_t$. After refinement procedure, we get refined proposals set $P'_{ssad}$, which is the final proposal results.

\subsection{Temporal Action Localization}

Since most videos in ActivityNet dataset only contain one action category, we use video-level action classification result as the category of temporal action proposals to get temporal action localization result.

\section{Experimental Results}

\subsection{Evaluation Metrics}

{\bf Localization.} In temporal action localization task, mean Average Precision (mAP) is used as the metric for evaluating result, which is similar with metrics used in object localization task. In detail, the official metric used in this task is the average mAP computed with tIoU thresholds  between 0.5 and 0.95 with the step size of 0.05.

{\bf Proposal.} In temporal action proposal task,  the area under the Average Recall vs. Average Number of Proposals per Video ({\bf AR-AN}) curve is used as the evaluation metric, where {\bf AR} is defined as the mean of all recall values using tIoU thresholds between 0.5 and 0.95 with a step size of 0.05. In this notebook, we call {\bf AR} with a certain number of  {\bf AN} as {\bf AR@AN}. For example, AR@100 means average recall with 100 proposals.

\subsection{Temporal Action Proposal}

\begin{figure}[tbp]
\centering
\begin{minipage}[b]{1.0\linewidth}
  \centering
  \centerline{\includegraphics[width=8.5cm]{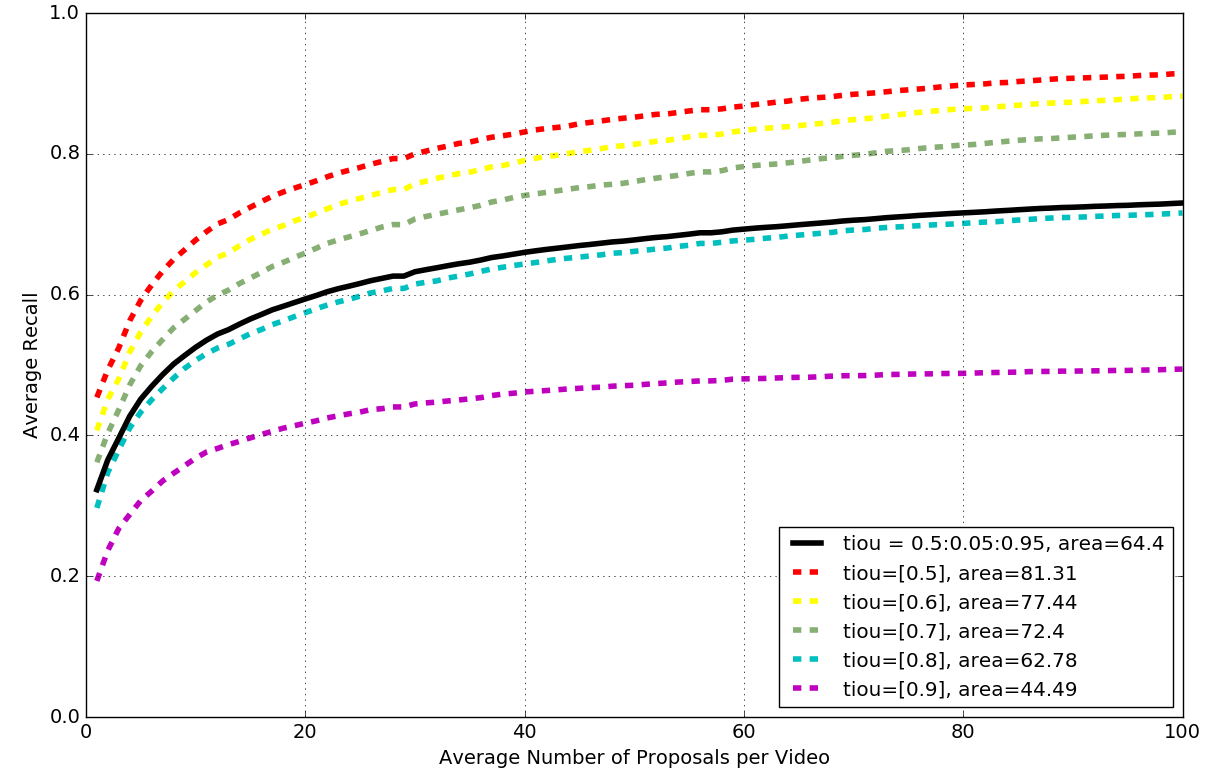}}
  \medskip
\end{minipage}
\caption{AR-AN curve of our proposal results in validation set. The area under black curve is the AR-AN score.
}
\label{result_proposal_img}
\end{figure}

The proposal performance on validation set of our approach are shown in Table \ref{result_proposal} and Figure \ref{result_proposal_img}. Our approach significantly outperform the baseline method and refined Prop-SSAD has better performance than Prop-SSAD. The boundaries refinement mainly improve the average recall with high tIoU. 

\begin{table}[!tbp]

\centering
\caption{Proposal Results on validation set of ActivityNet.}
\small
\begin{tabular}{p{3cm}<{\centering}ccc<{\centering}}
\toprule
Method  & AR@10 & AR@100  & AR-AN  \\
\midrule Uniform Random (baseline)  & 29.02 & 55.71 & 44.88 \\
Prop-SSAD  & 50.44 & 69.54 & 61.52 \\
Refined Prop-SSAD & 52.50 & 73.01 & 64.40 \\
\bottomrule
\end{tabular}
\label{result_proposal}
\normalsize
\end{table}

\subsection{Temporal Action Localization}

In the temporal action localization task, we directly use  proposals submitted in temporal action proposal task. For action categories, we use the video-level classification results of \cite{zhao2017cuhk}\footnote{Previously, we adopted classification results from result files of \cite{wang2016uts}. Recently we found that the classification accuracy of these results are unexpected high. Thus we replace it with classification results of \cite{zhao2017cuhk} and updated all related experiments accordingly.}. 

Evaluation results in validation set are shown in Table \ref{result_detection_val}. These results suggest that localization mAP  mainly depends on first several proposals. Therefore, we think AR-AN may not be the best evaluation metric for temporal action proposal task. AR with small proposals amount should has higher weight in evaluation metric.

Evaluation results in testing set are shown in Table \ref{result_detection_test}. Our approach significantly outperform other state-of-the-art approaches. We think the main contributor is our high quality temporal action proposals.

\begin{table}[tbp]

\centering
\caption{Action localization results on validation set. Results are evaluated by mAP with different IoU thresholds $\alpha$ and average mAP of IoU thresholds from 0.5 to 0.95. Ours@n means first n  proposals used for localization.}
\small
\begin{tabular}{p{2.2cm}<{\centering}|ccc|c<{\centering}}
\toprule
mAP  & 0.5  &  0.75  & 0.95  & Average mAP  \\
\hline Wang et al. \cite{wang2016uts}    & 42.28 & 3.76  & 0.05   & 14.85  \\
Shou et al. \cite{shou2017cdc}    & 43.83  & 25.88  & 0.21   & 22.77  \\
Xiong et. al. \cite{xiong2017pursuit}    & 39.12 & 23.48  & 5.49  & 23.98  \\
Ours@1   & 39.21  & 25.37  & 6.01   & 25.17  \\
Ours@5   & 42.57  & 28.26  & 6.83  & 27.86 \\
Ours@10 & 43.58   & 28.95  & 7.00  & 28.56 \\
Ours@25 & 44.14   & 29.42  & 7.07  & 28.96 \\
Ours@100 & 44.39   & 29.65  & 7.09  & 29.17 \\
\bottomrule
\end{tabular}
\label{result_detection_val}
\normalsize
\end{table}

\begin{table}[tbp]

\centering
\caption{Action localization results on testing set. Only average mAP is provided in evaluation server, which is calculated with IoU thresholds from 0.5 to 0.95.}
\small
\begin{tabular}{p{5cm}<{\centering}c<{\centering}}
\toprule
Method  & Average mAP    \\
\midrule Wang et. al. \cite{wang2016uts}    & 14.62 \\
Xiong et. al. \cite{xiong2017pursuit}    & 26.05 \\
Zhao et. al. \cite{zhao2017temporal}   & 28.28 \\
\midrule Ours result  & 32.26  \\
\bottomrule
\end{tabular}
\label{result_detection_test}
\normalsize
\end{table}

\section{Conclusion}

In this challenge, we mainly focus on the temporal action proposal task and obtains the salient performance in both temporal action proposal and temporal action localization task. Our results suggested that anchor mechanisms and temporal convolution can work well in temporal action proposal task. In the future, we will improve our framework such as training the whole networks end-to-end.

{\small
\bibliographystyle{ieee}
\bibliography{egbib}
}

\end{document}